\documentclass[letterpaper]{article} % DO NOT CHANGE THIS
\usepackage[draft]{aaai2027}  % DO NOT CHANGE THIS
\usepackage[hyphens]{url}  % DO NOT CHANGE THIS
\usepackage{graphicx} % DO NOT CHANGE THIS
\usepackage{natbib}  % DO NOT CHANGE THIS AND DO NOT ADD ANY OPTIONS TO IT
\usepackage{caption} % DO NOT CHANGE THIS AND DO NOT ADD ANY OPTIONS TO IT
\usepackage{algorithm}
\usepackage{algorithmic}

\usepackage{newfloat}
\usepackage{listings}
\DeclareCaptionStyle{ruled}{labelfont=normalfont,labelsep=colon,strut=off} % DO NOT CHANGE THIS
\floatstyle{ruled}
\newfloat{listing}{tb}{lst}{}
\floatname{listing}{Listing}

\usepackage{booktabs}

\usepackage{multirow}
\usepackage{amsmath}
\usepackage{amssymb}

\newcommand{\modelname}{VisKG-LM}
\newcommand{\rlpfull}{Relation-Labeled Paths}
\newcommand{\rlp}{RLP}   % abbreviation; use \rlpfull{} (\rlp{}) at first mention

\newcommand{\deepenc}{DeepEncoder}   % use as: \deepenc{} V2

\title{VisKG-LM: Compiling Knowledge Graphs into Visual Memory for Multiple-Choice Question Answering}
\author{
    Yixin Peng\textsuperscript{\rm 1}\equalcontrib\corresponding,
    Er Jin\textsuperscript{\rm 1}\equalcontrib,
    Shiwei Luo\textsuperscript{\rm 1}\equalcontrib,
    Diego Collarana\textsuperscript{\rm 2},
    Stefan Decker\textsuperscript{\rm 1,2}
}

\affiliations{
    \textsuperscript{\rm 1}RWTH Aachen University, Aachen, Germany\\
    \textsuperscript{\rm 2}Fraunhofer FIT, Sankt Augustin, Germany\\
    \{peng,jin,decker\}@dbis.rwth-aachen.de,
    shiwei.luo@rwth-aachen.de,\\
    diego.collarana.vargas@fit.fraunhofer.de
}

\begin{document}

\maketitle

\begin{abstract}
Knowledge graphs are usually integrated into question answering by encoding a
retrieved subgraph with a graph neural network and fusing it with the language
model in the online inference path. The same subgraph is therefore re-encoded
from scratch every time a pair is scored, across training epochs, seeds, and
evaluation runs, even though the knowledge graph never changes. We ask whether the retrieved knowledge graphs can instead be
compiled once, offline, and then accessed as read-only memory. \textbf{\modelname{}} shows that it can, by decoupling graph encoding from
language reasoning. It serializes each retrieved candidate-specific subgraph as
\rlpfull{} and renders the result as an image whose two-dimensional layout
preserves the branching structure of the paths. Each image is encoded once,
offline, and cached for reuse. At inference, the language model contextualizes the question and candidate
from text alone, and only its final layer consults the cached visual memory,
reading both its global layout and its local relational detail. The graph information thus
enters only after the text has been understood. On the test sets of
CommonsenseQA, OpenBookQA, and MedQA-USMLE, \modelname{} improves over
GreaseLM by $1.2$, $0.8$, and $4.3$ points, respectively, while matching or surpassing GraphVis, a $7$B vision-language model, with only about $400$M online parameters. Against a matched text-only control that
receives the identical \rlpfull{}, it gains $4.2$, $6.5$, and $5.1$ points across
the three benchmarks. These gains show that the complete visual-memory interface
adds value beyond path textualization alone and support compiled visual memory
as an alternative to online graph propagation.
\end{abstract}

\section{Introduction}
\label{sec:introduction}

Pretrained language models (LMs) encode substantial factual knowledge in their parameters and achieve strong performance across a wide range of downstream NLP tasks~\cite{roberts-etal-2020-much,petroni-etal-2019-language}. However, knowledge-intensive tasks such as multiple-choice question answering (MCQA) often require factual or domain-specific knowledge that is not reliably captured by parametric memory alone. This limitation is especially pronounced for smaller LMs, which generally show lower accuracy and less consistent predictions on challenging MCQA benchmarks~\cite{pinhanez2026small}. For instance, CommonsenseQA~\cite{talmor2019commonsenseqa}, OpenBookQA~\cite{mihaylov2018can}, and MedQA-USMLE~\cite{jin2021medqa} require commonsense associations, elementary scientific facts, and clinical knowledge, respectively. Recent studies therefore augment LMs with external knowledge, leading to substantial improvements in MCQA performance~\cite{ge2026expert,li-etal-2026-ur2,zou-wang-2026-iterative}. Much of this knowledge is relational, and knowledge graphs (KGs) provide a compact representation in which entities are modeled as nodes, relations as typed edges, and multi-hop facts as explicit paths~\cite{lewis2020retrieval,ji2021survey,hogan2021knowledge}. 
% Existing KG-enhanced MCQA methods commonly retrieve an answer-candidate-specific subgraph, encode it with a graph neural network (GNN), and fuse the resulting graph representation with the LM's textual representations~\cite{feng2020mhgrn,yasunaga2021qagnn,zhang2022greaselm}.

\begin{table}[t]
\centering
\begingroup
\small
\setlength{\tabcolsep}{1mm}
\begin{tabular*}{\columnwidth}{@{\extracolsep{\fill}}lccccr@{}}
\toprule
\textbf{Approach} &
\textbf{Topo.} &
\begin{tabular}[c]{@{}c@{}}\textbf{No}\\\textbf{G/V}\end{tabular} &
\textbf{Cache} &
\begin{tabular}[c]{@{}c@{}}\textbf{Late}\\\textbf{fusion}\end{tabular} &
\textbf{Params.} \\
\midrule
Pure-text LM
& \(\triangle\) & \(\checkmark\) & \(\times\) & \(\times\)
& 355M \\
LM--GNN
& \(\checkmark\) & \(\times\) & \(\times\) & \(\triangle\)
& 356--360M \\
LVLM
& \(\checkmark\) & \(\times\) & \(\times\) & \(\times\)
& 7B \\
\midrule
\textbf{\modelname{} (ours)}
& \(\checkmark\) & \(\checkmark\) & \(\checkmark\) & \(\checkmark\)
& 402M \\
\bottomrule
\end{tabular*}
\endgroup
\caption{\textbf{Baseline comparison.}
\(\checkmark\), \(\triangle\), and \(\times\) denote full, partial, and no support, respectively. 'Topo.': graph topology preservation; 'No G/V': no online graph/vision encoder; 'Cache': reusable encoded KGs; and 'Late fusion': graph information fusion only after text contextualized.}
\label{tab:interface-comparison}
\end{table}

The central challenge lies in determining how the structured information in a KG is exposed to the LM, when the graph representation is computed, and at which stage it is fused with the textual representation. Existing KG-enhanced MCQA methods can be broadly grouped into three paradigms: pure-text LM approaches, LM--GNN approaches, and large vision--language model (LVLM) approaches. As summarized in Table~\ref{tab:interface-comparison}, these paradigms make different trade-offs in terms of topology preservation, online graph or vision computation, representation reusability, fusion stage, and model scale. Pure-text LM approaches serialize retrieved graphs as textual sequences and process them with a text-only LM~\cite{jiang-etal-2023-structgpt,sun2024thinkongraph,edge2024local}. 
However, linearization obscures the original graph structure, making path order, shared prefixes, branching patterns, and structural connections no longer preserved. LM--GNN approaches, including KagNet~\cite{lin2019kagnet}, MHGRN~\cite{feng2020mhgrn}, QA-GNN~\cite{yasunaga2021qagnn}, and GreaseLM~\cite{zhang2022greaselm}, preserve graph topology while keeping the overall model at a moderate parameter scale. However, graph representations are commonly involved and fused in multiple stages of text--graph interaction, and it cannot be precomputed once and reliably reused across training epochs, random seeds, or evaluation runs. Lastly, LVLM approaches such as GraphVis~\cite{deng2024graphvis} render graphs as images to preserve their topological structure and extend graph information into the visual modality, allowing the language model to incorporate graph information through multimodal fusion. However, these approaches require training a 7B LVLM, resulting in substantially higher training and inference costs than those of pure-text LM and LM--GNN approaches. Moreover, because the visual graph representations are generated by a trainable vision-language encoder and evolve throughout training, they cannot be encoded once offline and reused as a fixed visual memory.

These limitations motivate a method that simultaneously preserves explicit graph topology, maintains a moderate parameter scale, and decouples graph encoding from language reasoning. We present \modelname{}, which addresses these limitations by compiling the
retrieved KGs once, offline, into a read-only visual memory.
Rather than linearizing a subgraph into a text sequence, \modelname{} converts each candidate-specific subgraph into
\rlpfull{} (\rlp{}) and renders it as a two-dimensional document image whose
layout explicitly preserves entity names, typed relations, path order, and orphan
nodes. This image is then encoded offline by a document encoder, \deepenc{} V2~\cite{wei2026deepseekocr2},
into cached visual representations, so the visual encoding is computed only once
and can be reused across training epochs, random seeds, and evaluation runs. At inference time, the lower layers of the LM process only the question and answer
candidate in text form, while graph information is introduced only at the top layer.
Specifically, a gated cross-attention module is applied at the top layer, allowing
the model to start from the original text-only representation and gradually learn
how much graph information to incorporate. Finally, a dual-view readout summarizes
the cached visual memory at both the global and local levels. Our contributions are as follows:
\begin{itemize}
    \item We propose a KG-enhanced MCQA architecture that replaces online GNN
    message passing with an offline-compiled, read-only visual memory, so each retrieved subgraph is encoded once and resuable for later stage. 
    \item We introduce a strict late-fusion design in which lower LM layers contextualize the question and candidate without graph access, and only the top layer queries the cached memory via unidirectional gated cross-attention; a dual-view summarizer then reads its global-layout and local-relational views for scoring. 
    \item Across CSQA, OBQA, and MedQA-USMLE, \modelname{} consistently outperforms strong LM--GNN baselines, while matching or surpassing GraphVis, a $7$B vision-language model, with only about $400$M online parameters. Compared with a matched text-only control provided with exactly the same paths, \modelname{} yields absolute gains of $4.2$, $6.5$, and $5.1$ percentage points on the three benchmarks, respectively, demonstrating that its improvements extend beyond path textualization alone.
\end{itemize}

\section{Related Work}
\label{sec:related}

\paragraph{KG-Enhanced Language Models.}

KG-enhanced QA methods differ primarily in where and how structured graph information enters the language model. 
LLM-centric approaches convert retrieved subgraphs into text using serialized paths, triples, or textual summaries and process them with a text-only LM~\cite{jiang-etal-2023-structgpt,sun2024thinkongraph,edge2024local,luo2024rog,he2024gretriever,wan2025digest}. Linearizing a graph weakens its explicit structure, path order may remain recoverable from the sequence, but shared prefixes, branching and cross-path connections are no longer directly represented.
Moreover, the serialized evidence must be processed again by the LM for every query. 
LM–GNN approaches tightly couple the text and graph encoders, scoring each question–candidate pair by jointly processing its retrieved subgraph and textual context.
MHGRN~\cite{feng2020mhgrn} performs MCQA statement-conditioned message passing over relational paths of up to \textit{k} hops.
QA-GNN~\cite{yasunaga2021qagnn} introduces the LM-encoded question--answer
context as an additional node in a joint graph and updates it together with
retrieved subgraph nodes. GreaseLM~\cite{zhang2022greaselm} interleaves upper
LM blocks with GNN layers, enabling bidirectional information exchange across multiple
layers. DRAGON~\cite{yasunaga2022dragon} further pretrains a GreaseLM-style
bidirectional text--graph model using masked language modeling and KG link prediction. 
JointLK~\cite{sun2022jointlk} applies dense bidirectional attention between
language tokens and KG nodes and recursively prunes the graph. QAT~\cite{park2023qat} replaces explicit GNN message passing with
relation-centric meta-path tokens. These approaches preserve graph structure and support task-adaptive reasoning, but keep the graph encoder in the loop at both training and inference, so its output cannot be compiled once and reused.

\paragraph{Visual Approaches for Graph Information.}

Linearizing a graph as a flat sequence of triples can make path order, shared prefixes, branching structure, and local grouping difficult to recover. GraphVis~\cite{deng2024graphvis} instead renders retrieved subgraph as node--link diagram and uses a two-stage procedure to fine-tune a 7B LVLM (LLaVA-v1.6-Mistral~\cite{liu2024improved}). Similarly, GITA~\cite{wei2024gita} combines visual graph
renderings with textual graph descriptions for general graph reasoning. VGCure~\cite{zhu-etal-2025-benchmarking} also show that graph-structure-aware self-supervised training can improve the ability of LVLMs to understand complex relational and structural information in graph images. Although these methods demonstrate the value of visualizing graph structure, they typically require computationally expensive LVLM training or fine-tuning. BLIP-2~\cite{li2023blip2} suggests a more parameter-efficient alternative: rather than jointly updating large vision and language backbones, it keeps both backbones frozen and connects them through a lightweight query transformer. Following this principle, \modelname{} renders each candidate-specific subgraph as a path-ordered visual document using \rlp{}, preserving graph structure without the node and edge overlaps common in compact node--link diagrams. A frozen vision encoder processes each document once, and the resulting visual tokens are cached for reuse. At the top layer, after the question and candidate have been contextualized, the dual-view memory summarizer extracts global-layout and local-relational summaries for answer scoring.

\paragraph{Document-Oriented Visual Encoders.}

Donut~\cite{kim2022donut} and Nougat~\cite{blecher2024nougat}
introduce an OCR-free paradigm for document understanding, and subsequent models further emphasize two features: faithful preservation of document structure and compact visual-token sequences~\cite{lee2023pix2struct,hu2024mplugdocowl15,hu2024mplugdocowl2}. DeepSeek-OCR~\cite{wei2025deepseek} pushes it further, compressing a high-resolution document into a short token sequence via its DeepEncoder. DeepSeek-OCR 2~\cite{wei2026deepseekocr2} retains this compressed visual tokenizer but replaces the CLIP-based component with an LM-style vision encoder, letting the compressed tokens exchange information through bidirectional attention while causal-flow queries reorganize the document information. Both features, structure-faithful representation and compact, cacheable token sequence, are well suited to our \rlp{} renderings, in which the key information is carried by entity names, typed relations, path order and branch symbols. We therefore use \deepenc{} V2 as a frozen, offline encoder that turns each rendered subgraph into cached, read-only visual memory.

\section{Method}
\label{sec:method}

\begin{figure}[t]
    \centering
    \includegraphics[width=\columnwidth]{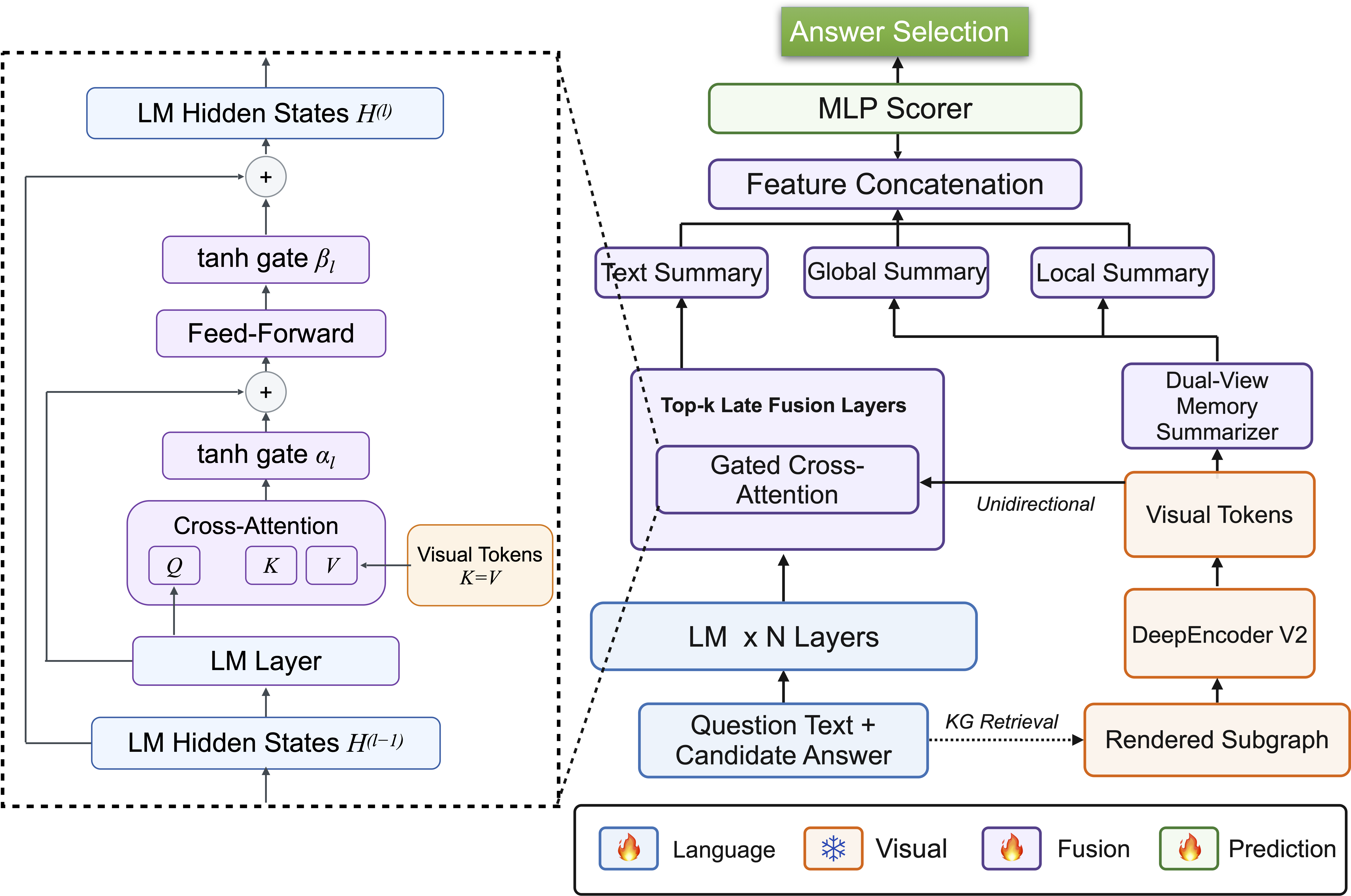}
    \caption{\textbf{The architecture of \modelname{}.}
    For each question--candidate pair, the retrieved subgraph is rendered and
    encoded offline (orange). The LM first contextualizes the text independently, after which the cached visual tokens are fused unidirectionally into the top $k$ layers. In each fusion layer, visual tokens serve as K and V for
    gated cross-attention, with separate zero-initialized $\tanh$ gates controlling the attention and FFN residuals. The final scorer concatenates the text summary
    with independently learned global and local summaries to rank the answer choices.}
    \label{fig:architecture}
\end{figure}

\paragraph{Problem Setup and Overview.}

Let $\mathcal{D}=\{(c_i,q_i,\mathcal{A}_i,y_i)\}$ be an MCQA dataset, where
$c_i$ is optional context, $q_i$ is a question,
$\mathcal{A}_i=\{a_{i,1},\ldots,a_{i,C_i}\}$ is the candidate set, where $C_i=5$ for CSQA and $C_i=4$ for OBQA, MedQA, and $y_i\in\{1,\ldots,C_i\}$ is the ground-truth index. The task is to select the correct candidate: the model assigns a score
$s_{i,j}$ to each triple $(c_i,q_i,a_{i,j})$ and normalizes these scores over the
$C_i$ candidates of question $i$ (Eq.~\ref{eq:score}). To ground each candidate in
external knowledge, we reuse the entity-linking subgraph retriever based on
QA-GNN, which maps each pair
$(q_i,a_{i,j})$ to a subgraph $\mathcal{G}^{(i,j)}_{\mathrm{sub}}$ of a
KG $\mathcal{G}=(\mathcal{V},\mathcal{E},\mathcal{R})$ with typed directed edges
$\mathcal{E}\subseteq\mathcal{V}\times\mathcal{R}\times\mathcal{V}$.

Figure~\ref{fig:architecture} illustrates \modelname{}. Unlike QA-GNN and
GreaseLM, which encode $\mathcal{G}^{(i,j)}_{\mathrm{sub}}$ online with a GNN coupled to the LM,
\modelname{} renders and encodes each subgraph offline as cached visual tokens
$\mathcal{V}_{i,j}=[\mathcal{V}^{g}_{i,j};\mathcal{V}^{l}_{i,j}]$, which
concatenate global and local views. A trainable projector maps the normalized visual 
tokens into the LM hidden space, yielding
$V'_{i,j}=\operatorname{Proj}(\operatorname{LN}(\mathcal{V}_{i,j}))\in
\mathbb{R}^{M\times d_\ell}$, where $M$ is the total number of cached visual
tokens, $\operatorname{LN}$ denotes layer normalization, and $d_\ell$ is the LM
hidden size.

For the language branch, the text
$X_{i,j}=\operatorname{Concat}(c_i,q_i,a_{i,j})$ is embedded into $H^0_{i,j}\in\mathbb{R}^{S\times d_\ell}$, where $S$ is the text sequence length. 

Let the LM has n layers, of which the top k perform fusuion. Each layer then update the hidden.

\begin{equation}
H^{\ell}_{i,j}=\begin{cases}
f_{\ell}(H^{\ell-1}_{i,j}), & \ell\leq N-k,\\
\operatorname{Fuse}_{\ell}(H^{\ell-1}_{i,j},V'_{i,j}), & \ell>N-k,
\end{cases}
\end{equation}
where $f_\ell$ is a pretrained LM layer, and $\operatorname{Fuse}_{\ell}$ injects the visual tokens $V'_{i,j}$ through gated cross-attention.  

\paragraph{Visual Knowledge Memory.}
\label{subsec:visual-memory}

\begin{figure*}[t]
    \centering
    \includegraphics[width=\textwidth]{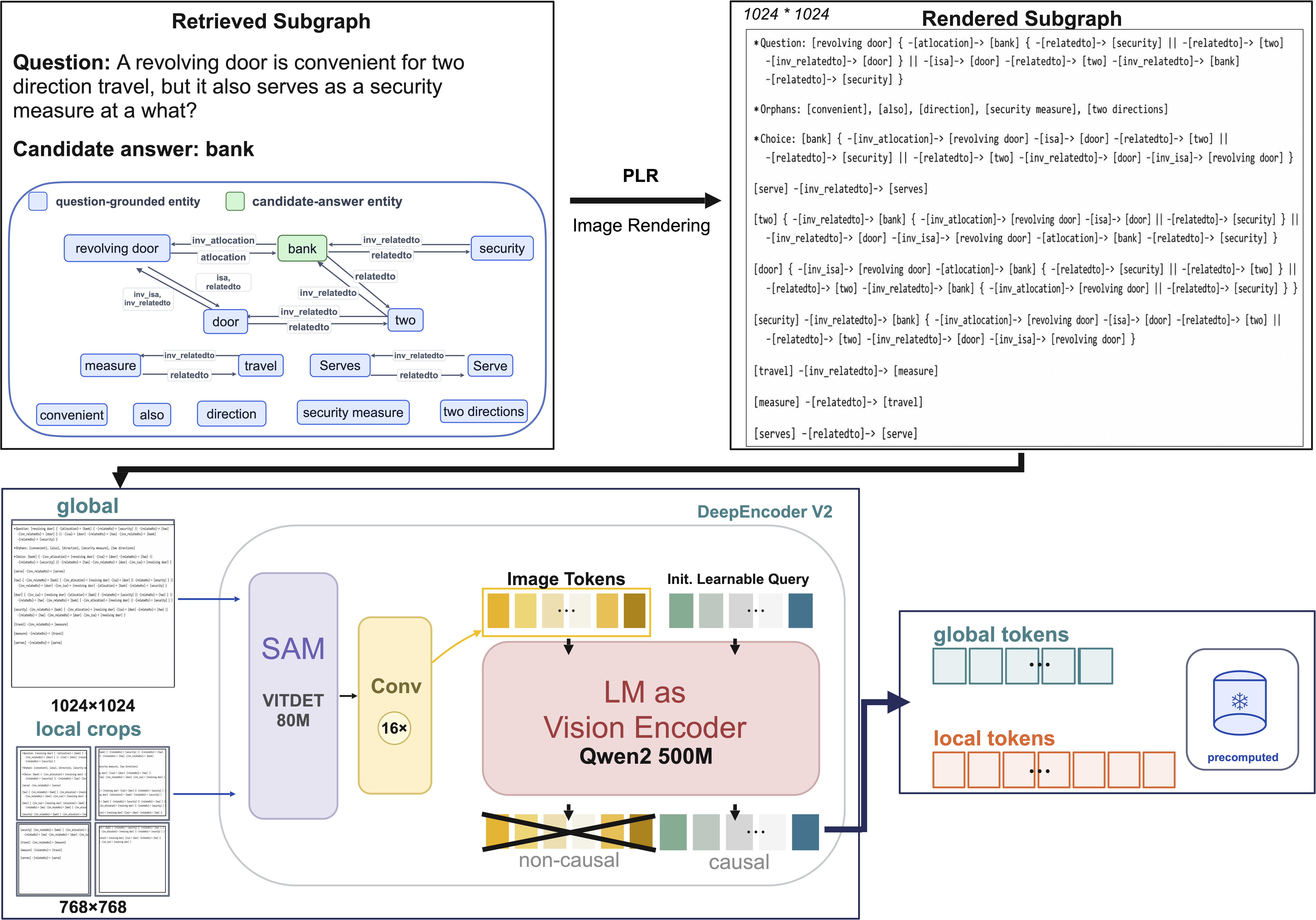}
    \caption{\textbf{Offline construction of candidate-specific visual
    memory.} \emph{Top left:} The retrieved subgraph for an example
    question--candidate pair, with question-grounded and candidate-answer
    entities highlighted. \emph{Top right:} \rlp{} serialization reorganizes
    the subgraph's nodes and relations into a structured text document, which
    is then rendered on a $1024{\times}1024$ canvas. \emph{Bottom:} \deepenc{} V2
    processes the full image as a global view and four $768{\times}768$ crops
    as local views. The final non-causal visual tokens are discarded, and only the
    final causal-query tokens are cached: $256$ global tokens and
    $4{\times}144$ local tokens.}
    \label{fig:visual_memory_construction}
\end{figure*}

The offline pipeline constructs candidate-specific visual memory as three steps.
First, \rlp{} serialization reorganizes the nodes and relations in the retrieved
subgraph $\mathcal{G}^{(i,j)}_{\mathrm{sub}}$ into a structured text document
$\mathcal{T}^{(i,j)}$. The document is then rendered on a
$1024{\times}1024$ canvas to produce an image
$I^{(i,j)}_{\mathrm{KG}}$. Finally, \deepenc{} V2 encodes global and local
views of this image into cached visual tokens $\mathcal{V}_{i,j}$.
Figure~\ref{fig:visual_memory_construction} shows the complete process.

The \rlp{} serializer reorganizes the retrieved nodes and relations into structured text. As shown in
Figure~\ref{fig:visual_memory_construction}, shared path prefixes are merged,
and question-seed, answer-seed, and orphan evidence is placed in separate
blocks. The serializer gives priority to paths that connect question and
answer entities, labels reverse traversals with inverse relations, and retains
parallel paths. To bound the search cost, it keeps at most a fixed number of
paths per seed entity. Shorter paths are selected first, and paths of the same
length are ordered alphabetically by their entity sequences, producing the
same document across runs. The renderer then lays out $\mathcal{T}^{(i,j)}$ in reading order on a fixed
$1024{\times}1024$ canvas. Each entity or relation is treated as one layout
unit and is never split across lines. The resulting document image is denoted
by $I^{(i,j)}_{\mathrm{KG}}$. We then encode $I^{(i,j)}_{\mathrm{KG}}$ with a frozen \deepenc{} V2. Its preprocessing keeps the complete
$1024{\times}1024$ image as the global view. To obtain the local views, it
resizes the image to $1536{\times}1536$ and divides it into a $2{\times}2$
grid of four non-overlapping $768{\times}768$ crops. The global view preserves
the overall document layout, while the local views make small entity and
relation strings easier to encode. We discard the non-causal visual tokens and
cache only the final causal-query tokens. The global view yields $256$ tokens,
and the four local views yield $144$ tokens each, giving
$M=256+4\times144=832$ cached tokens per candidate. Thus,
$\mathcal{V}^{g}_{i,j}\in\mathbb{R}^{256\times d_v}$ and
$\mathcal{V}^{l}_{i,j}\in\mathbb{R}^{576\times d_v}$, where $d_v$ is the
output dimension of \deepenc{} V2.

\paragraph{Late Gated Cross-Attention.}
\label{subsec:gated-fusion}

At each fusion layer, the mapped visual tokens $V'_{i,j}$ serve as a fixed
key--value memory. Cross-attention is therefore unidirectional: the language
representations attend to the visual memory, while the visual tokens remain
independent of the text. The LM layer first contextualizes the current language
representations, after which cross-attention injects candidate-specific graph
information. For question $i$ and candidate $j$, the layer updates are

\begin{subequations}\label{eq:visual-fusion}
\begin{align}
\bar H_{i,j}^{\ell}
&= f_\ell\!\left(H_{i,j}^{\ell-1}\right),
\label{eq:visual-fusion-lm}\\
U_{i,j}^{\ell}
&= \operatorname{CrossAttn}\!\left(
    \operatorname{LN}\!\left(\bar H_{i,j}^{\ell}\right),
    V'_{i,j},
    V'_{i,j}
\right),
\label{eq:visual-fusion-attn}\\
\widetilde H_{i,j}^{\ell}
&= \bar H_{i,j}^{\ell}
   + \tanh\!\left(\alpha_\ell\right) U_{i,j}^{\ell},
\label{eq:visual-fusion-attn-gate}\\
H_{i,j}^{\ell}
&= \widetilde H_{i,j}^{\ell}
   + \tanh\!\left(\beta_\ell\right)
     \operatorname{FFN}\!\left(
       \operatorname{LN}\!\left(\widetilde H_{i,j}^{\ell}\right)
     \right).
\label{eq:visual-fusion-ffn}
\end{align}
\end{subequations}

Here, $\bar H_{i,j}^{\ell}\in\mathbb{R}^{S\times d_\ell}$ is the output of
the ${\ell-1}$ LM layer, $U_{i,j}^{\ell}\in\mathbb{R}^{S\times d_\ell}$
is the visual information returned by cross-attention, and $\operatorname{CrossAttn}$ and $\operatorname{LN}$ denote cross-attention and layer normalization. The scalar gates
$\alpha_\ell$ and $\beta_\ell$ are shared across examples, independently
parameterized, and initialized to zero. This separation allows the
cross-attention and fusion-specific feed-forward residuals to learn different
contribution magnitudes.

% The effect of late gated fusion is evaluated in
% Table~\ref{tab:component-ablations}.

\paragraph{Dual-View Summarization and Scoring.}
\label{subsec:dvms}

DVMS summarizes the global and local visual tokens separately, which employs a small bank of
learnable queries for each view to selectively aggregate the most relevant visual information. For branch $b\in\{g,l\}$, let
$V_{i,j}^{\prime b}\in\mathbb{R}^{M_b\times d_\ell}$ denote the corresponding
partition of $V'_{i,j}$, where $M_g+M_l=M$. Each branch maintains an
independent learnable query bank
$Q_b\in\mathbb{R}^{L_b\times d_\ell}$, where $L_g=G$ and $L_l=L$ denote the
numbers of global and local queries, respectively. The query banks $Q_g$ and
$Q_l$ are separately parameterized and shared across all
question--candidate pairs. For notational convenience, we define
\begin{equation}
\bar Q_b=\operatorname{LN}(Q_b),
\qquad
\bar V_{i,j}^{b}=\operatorname{LN}(V_{i,j}^{\prime b}).
\end{equation}
The branch-specific visual summaries are then computed as
\begin{subequations}\label{eq:dvms}
\begin{align}
\widetilde Z_{i,j}^{b}
&=
\bar Q_b
+
\operatorname{CrossAttn}\!\left(
\bar Q_b,\,
\bar V_{i,j}^{b},\,
\bar V_{i,j}^{b}
\right),
\label{eq:dvms-attn}\\
Z_{i,j}^{b}
&=
\operatorname{LN}\!\left(
\widetilde Z_{i,j}^{b}
+
\operatorname{FFN}\!\left(
\operatorname{LN}(\widetilde Z_{i,j}^{b})
\right)
\right),
\label{eq:dvms-ffn}\\
h_{i,j}^{b}
&=
\frac{1}{L_b}
\sum_{t=1}^{L_b} Z_{i,j,t}^{b},
\label{eq:dvms-pool}
\end{align}
\end{subequations}
where, $\operatorname{FFN}$ denotes a feed-forward network, and $Z_{i,j,t}^b$ is the output of
the $t$-th query. The $L_b$ query outputs are averaged to produce a single
summary token $h_{i,j}^b\in\mathbb{R}^{d_\ell}$ for branch $b$. For each question--candidate pair, the scorer concatenates the LM's final \texttt{[CLS]}
representation $h_{i,j}^{\mathrm{text}}$ with the global and local summaries,
\begin{equation}
\label{eq:score}
h_{i,j}=[h_{i,j}^{\mathrm{text}}\|h_{i,j}^g\|h_{i,j}^l]
\in\mathbb{R}^{3d_\ell}, \; s_{i,j}=\operatorname{MLP}(h_{i,j}).
\end{equation} 
The model is trained using a cross-entropy loss over the candidate scores to
assign the highest score to the correct answer.

\section{Experiments and Results}
\label{sec:experiments}

\paragraph{Datasets and Protocol.}

We evaluate all baselines and model variants on CSQA, OBQA, and MedQA-USMLE using the data splits and retrieved KGs provided by GreaseLM. CSQA and OBQA use ConceptNet, whereas MedQA uses the Disease Database and DrugBank. For CSQA, we follow the in-house (IH) evaluation protocol of KagNet~\cite{lin2019kagnet}, reporting results on IHdev and the held-out IHtest split. OBQA and MedQA use their standard splits. Table~\ref{tab:dataset-statistics} summarizes the split sizes and the number of answer candidates per question. Appendix~\ref{supp:dataset-statistics} provides the official CSQA split sizes and the number of candidate-level images generated during preprocessing.

\begin{table}[t]
\centering
\begingroup
\small
\setlength{\tabcolsep}{3pt}
\begin{tabular}{@{}lrrrr@{}}
\toprule
\textbf{Dataset} & \textbf{Choices} & \textbf{Train} & \textbf{Dev} &
\textbf{Test} \\
\midrule
CSQA (IH) & 5 & 8,500 & 1,221 & 1,241 \\
OBQA & 4 & 4,957 & 500 & 500 \\
MedQA-USMLE & 4 & 10,178 & 1,272 & 1,273 \\
\bottomrule
\end{tabular}
\endgroup
\caption{\textbf{Datasets statistics.} 
CSQA follows the five-choice in-house (IH) splits, whereas four-choice OBQA
and MedQA-USMLE use their standard splits~\cite{zhang2022greaselm}.}
\label{tab:dataset-statistics}
\end{table}

\paragraph{Models and Training.} We use RoBERTa-large~\cite{liu2019roberta} as the text backbone for CSQA and
OBQA, and SapBERT-Base~\cite{liu2021sapbert} for MedQA\@. The default offline
visual encoder is \deepenc{} V2. To test sensitivity to the choice of visual
encoder, we replace it with ViT-Base (ViT-B) or \deepenc{} V1, while keeping the renderer, projector,
fusion module, and text backbone fixed. Each fusion block has eight attention
heads, and the default fusion depth is $k=1$. DVMS uses $(G,L)=(4,8)$ for CSQA
and MedQA and $(G,L)=(1,8)$ for OBQA. We optimize all models with AdamW, using weight decay of $0.03$, batch size of $256$, and a $300$-step warm-up followed by a constant learning
rate. The backbone learning rate is $10^{-5}$ for CSQA and OBQA and
$5\times10^{-5}$ for MedQA; the projector, cross-attention module, DVMS, and
scorer all use $10^{-4}$. We train for up to $40$, $80$, and $70$ epochs on
CSQA, OBQA, and MedQA, respectively. The text backbone is frozen for the first
four epochs and then trained jointly with the rest of the model. Training uses
bfloat16 on a single 80GB NVIDIA H100 GPU\@. Reported $\pm$ values are
standard deviations over nine runs, with three runs for each seed in
$\{5,6,7\}$. Appendix~\ref{supp:implementation} lists the remaining implementation details.

\paragraph{Baselines}

We compare against KG-free base LMs, MHGRN, QA-GNN, GreaseLM and GraphVis.
\deepenc{} V2 and the other visual encoders run only offline to build the fixed visual cache, so they add no parameters to \modelname{}'s online model, as shown in Table~\ref{tab:main-results}, which stays far smaller than the $7$B GraphVis. As our main control, we give the base LMs the same candidate-specific
\rlp{}, but present it directly as text. The control neither renders the \rlp{} as an image nor processes it with a visual encoder or fuses
it with visual memory. Therefore, the performance difference isolates the effect of our visual encoding and fusion architecture. LM-GNN baselines result on OBQA use AristoRoBERTa~\cite{clark2019aristo}, whereas our models use vanilla RoBERTa-large. AristoRoBERTa is based on
RoBERTa-large but additionally incorporates question-specific facts curated by
\cite{clark2019aristo}; it reportedly outperforms vanilla RoBERTa-large on OBQA.

\begin{table*}[t]
\centering
\begingroup
\small
\setlength{\tabcolsep}{2pt}
\begin{tabular*}{\textwidth}{@{\extracolsep{\fill}}llccccc@{}}
\toprule
\multirow{2}{*}{\textbf{Cat.}} &
\multirow{2}{*}{\textbf{Method}} &
\textbf{Params.} &
\multicolumn{2}{c}{\textbf{CSQA}} &
\textbf{OBQA} & \textbf{MedQA} \\
\cmidrule(lr){3-3}\cmidrule(lr){4-5}\cmidrule(lr){6-6}\cmidrule(l){7-7}
& & \textit{CSQA/OBQA; MedQA} & \textbf{IHdev} & \textbf{IHtest} &
\textbf{Test} & \textbf{Test} \\
\midrule
\multirow{5}{*}{\textit{Prior}} &
Base LM (w/o KG)\textsuperscript{\dag}
& 355M; 109.5M & $73.1\pm0.5$ & $68.7\pm0.6$ & $78.4\pm0.6$ & $37.2\pm0.1$ \\
& MHGRN~\cite{feng2020mhgrn}\textsuperscript{\dag}
& 356M; 113M & $74.5\pm0.1$ & $71.1\pm0.8$ & $80.6\pm0.2$ & $37.5\pm0.1$ \\
& QA-GNN~\cite{yasunaga2021qagnn}\textsuperscript{\dag}
& 360M; 114M & $76.5\pm0.2$ & $73.4\pm0.9$ & $82.8\pm0.7$ & $38.0\pm0.5$ \\
& GreaseLM~\cite{zhang2022greaselm}\textsuperscript{\dag}
& 359M; 119.6M & $78.0\pm0.6$ & $74.0\pm0.5$ & $84.8\pm0.9$ & $38.4\pm0.2$ \\
& GraphVis (7B LVLM)~\cite{deng2024graphvis}
& 7B; 7B & $\mathbf{80.3\pm0.6}$ & $75.1\pm0.9$ & $85.5\pm0.1$ & $40.6\pm0.2$ \\
\midrule
\textit{Text-only} &
Base LM with \rlp{} (text only)
& 355M; 109.5M & $77.6\pm0.2$ & $71.0\pm0.3$ & $79.1\pm0.4$ & $37.6\pm0.1$ \\
\midrule
\multirow{3}{*}{\textit{Ours}} &
\modelname{} + ViT-B
& 401.4M; 137.7M & $75.3\pm0.2$ & $71.0\pm0.6$ & $82.0\pm0.1$ & $40.7\pm0.3$ \\
& \modelname{} + \deepenc{} V1
& 404.1M; 139.8M & $75.1\pm0.7$ & $72.0\pm0.9$ & $81.4\pm0.1$ & $40.0\pm0.2$ \\
& \textbf{\modelname{} + \deepenc{} V2}
& 402.1M; 138.3M & $78.1\pm0.8$ & $\mathbf{75.2\pm0.4}$ &
$\mathbf{85.6\pm0.2}$ & $\mathbf{42.7\pm0.3}$ \\
\bottomrule
\end{tabular*}
\endgroup
\caption{\textbf{Main accuracy results (\%).} \textsuperscript{\dag}Prior CSQA,
OBQA, and MedQA results use RoBERTa-large, AristoRoBERTa, and SapBERT-Base,
respectively. Boldface marks the best available score in each accuracy column.}
\label{tab:main-results}
\end{table*}

\paragraph{Main Results.}

Table~\ref{tab:main-results} shows three main findings. First, \modelname{} outperforms the matched text-only control on all three benchmarks, by $4.2$, $6.5$, and $5.1$ points on CSQA IHtest, OBQA, and MedQA; since the control receives the identical \rlp{} as raw text, the gain reflects the visual-memory interface rather than path construction alone. Second, it also exceeds every reported LM--GNN baseline, most clearly on MedQA, where it reaches $42.7\%$, $4.3$ points above GreaseLM, and $5.2$ and $4.7$ points over MHGRN and QA-GNN, respectively. Third, despite using only $402.1$M online parameters, \modelname{} slightly outperforms the $7$B GraphVis on CSQA IHtest ($75.2\%$ vs.\ $75.1\%$) and OBQA ($85.6\%$ vs.\ $85.5\%$), while establishing a clearer advantage on MedQA ($42.7\%$ vs.\ $40.6\%$). GraphVis remains stronger only on CSQA IHdev. Overall, \modelname{} achieves competitive or superior performance to the much larger LVLM with approximately $18\times$ fewer online parameters. Replacing \deepenc{} V2 with ViT-B decreases the accuracy by $4.3$, $3.6$, and $2.0$ points on CSQA IHtest, OBQA, and MedQA, respectively, and \deepenc{} V1
reduces it by $3.2$, $4.2$, $2.7$ points. These drops suggest that document-oriented
compression benefits text-dense renderings. Both alternatives still outperform the
text-only control on all benchmarks. Thus, the visual-memory interface also works with encoders other than \deepenc{} V2.

\paragraph{Ablation Studies.}

\begin{table*}[t]
\centering
\begingroup
\def\abref#1{\textbf{#1}\,{\small(\textit{ref.})}}
\def\abdrop#1#2{#1\,{\small(\(-#2\))}}
\setlength{\tabcolsep}{5pt}
\begin{tabular*}{\textwidth}{@{\extracolsep{\fill}}lccc@{}}
\toprule
Ablation &
\multicolumn{3}{c}{Test accuracy (\%) (\(\Delta\))} \\
\cmidrule(l){2-4}
& CSQA & OBQA & MedQA \\
\midrule
Reference: full model (DVMS, $\tanh$ gating, $k=1$) &
\abref{75.2} & \abref{85.6} & \abref{42.7} \\
Replace DVMS readout with mean pooling &
\abdrop{73.9}{1.3} & \abdrop{83.6}{2.0} & \abdrop{40.0}{2.7} \\
Remove $\tanh$ gating &
\abdrop{73.6}{1.6} & \abdrop{84.4}{1.2} & \abdrop{39.8}{2.9} \\
\midrule
Deepen fusion layers (DVMS, $\tanh$ gating, $k:1\!\to\!2$) &
\abdrop{73.9}{1.3} & \abdrop{83.0}{2.6} & \abdrop{39.6}{3.1} \\
Deepen fusion layers (DVMS, $\tanh$ gating, $k:1\!\to\!4$) &
\abdrop{74.5}{0.7} & \abdrop{83.4}{2.2} & \abdrop{39.4}{3.3} \\
Deepen fusion layers (DVMS, $\tanh$ gating, $k:1\!\to\!6$) &
\abdrop{73.9}{1.3} & \abdrop{84.2}{1.4} & \abdrop{39.8}{2.9} \\
\bottomrule
\end{tabular*}
\endgroup
\caption{Component ablations. The full model uses
dual-view memory summarization (DVMS), $\tanh$-gated residuals, and memory
fusion only in the final LM layer ($k=1$). Each row changes only the named
component; parenthesized values are absolute percentage-point changes
from the full model.}
\label{tab:component-ablations}
\end{table*}

Our method is built around three design choices: DVMS summarizes the cached visual tokens using learnable queries, $\tanh$ gates regulate the amount of visual information injected into the LM, and visual memory is fused only in the final LM layer. Table~\ref{tab:component-ablations} evaluates these choices by replacing DVMS with mean pooling, removing the gates, and extending fusion to additional LM layers. Each modification consistently reduces accuracy across all three datasets, confirming that all three components contribute to the final performance. The degradation is particularly pronounced on MedQA, where the variants shown in the table incur drops of $2.7$--$3.3$ percentage points; the smaller but consistent decreases on CSQA and OBQA further support the effectiveness of these design choices. Appendices~\ref{supp:fusion-depth} and~\ref{supp:dvms-ablations} report the complete fusion-depth sweep and additional DVMS ablations, respectively.

\paragraph{Graph Representation and Corruption.}

To identify which components of \rlp{} drive performance, we compare relation semantics, structural marks, memory views, and canvas density in Table~\ref{tab:rendering-diagnostics}. The full \rlp{} representation performs best, reaching $42.66\%$ accuracy on MedQA. Removing relation labels reduces accuracy to $40.69\%$, even when the path and branch marks are retained, while entity groups without either relations or structural marks achieve $41.08\%$. This indicates that the main benefit comes from explicitly showing how entities are related, rather than from the structural marks alone. The local memory view is also important: using only the global view lowers the entity-group result from $41.08\%$ to $28.31\%$, showing that global layout alone loses substantial fine-grained information. Finally, reducing the global-only canvas from $1024^2$ to $512^2$ slightly improves accuracy to $29.77\%$ because the same content is packed more densely. Overall, the results show that explicit relation labels and the combination of global and local views are the key factors behind the effectiveness of \rlp{}.

\begin{table}[t]
\centering
\begingroup
\small
\setlength{\tabcolsep}{2pt}
\renewcommand{\arraystretch}{1.12}
\begin{tabular}{@{}lcccc@{}}
\toprule
\textbf{Representation} & \shortstack{\textbf{Relation}\\\textbf{labels}} &
\textbf{Marks} & \textbf{Memory} & \shortstack{\textbf{Test Acc.}\\\textbf{(\%)}} \\
\midrule
\rlp{} & Yes & Yes & $1024^2$, G+L & \textbf{42.66} \\
Entity groups only & No & No & $1024^2$, G+L & 41.08 \\
\shortstack[l]{Relation-free paths} & No & Yes &
$1024^2$, G+L & 40.69 \\
\shortstack[l]{Entity groups only} & No & No & $1024^2$, G &
28.31 \\
\shortstack[l]{Entity groups only} & No & No & $512^2$, G &
29.77 \\
\bottomrule
\end{tabular}
\endgroup
\caption{\textbf{Effect of rendering information on MedQA.}
``Marks'' are the path-order markers used in \rlp{}; G and L
denote global and local views. 
Reducing the canvas to $512^2$ retains all content while producing a denser layout with less whitespace.}
\label{tab:rendering-diagnostics}
\end{table}

We next perturb the graph structure by replacing relation labels, deleting
edges, or corrupting entity names other than the question and answer seeds.
For each corruption operator and ratio
($\rho\in{0,25,50,75,100\%}$), both models are trained and evaluated under
the same corruption setting. \modelname{} consistently outperforms GreaseLM
across all settings, with a margin of at least $3.8$ percentage points.
Compared with its corresponding clean result, \modelname{} loses at most
$0.45$ points. GreaseLM, however, is similarly insensitive to increasing
corruption. Thus, rather than demonstrating robustness to unseen graph
perturbations, these results show that \modelname{}'s performance advantage
persists even when training and evaluation are conducted on increasingly
corrupted graphs. Table~\ref{tab:corruption-endpoints} reports the clean and
fully corrupted endpoints, while Appendix~\ref{supp:graph-corruption} provides
the intermediate corruption ratios and visual examples of the three
operators.

\begin{table}[t]
\centering
\begingroup
\small
\setlength{\tabcolsep}{3pt}
\begin{tabular}{@{}lcccc@{}}
\toprule
\textbf{Model} & \textbf{Clean} & \shortstack{\textbf{Edge}\\\textbf{label}} &
\shortstack{\textbf{Edge}\\\textbf{deletion}} & \textbf{Entity} \\
\midrule
GreaseLM & 38.41 & 38.49 & 38.18 & 38.41 \\
\textbf{\modelname{}} & \textbf{42.66} & \textbf{42.34} &
\textbf{42.58} & \textbf{42.41} \\
\bottomrule
\end{tabular}
\endgroup
\caption{\textbf{MedQA accuracy (\%) under matched train--test graph
corruption.} Clean denotes $\rho=0$, while the remaining columns report
$\rho=100\%$ corruption through relation-label replacement, edge deletion, or
non-seed entity replacement.}
\label{tab:corruption-endpoints}
\end{table}

\paragraph{Online Cost.}
We compare the online efficiency of \modelname{} and GreaseLM on MedQA. As shown in Table~\ref{tab:online-cost},
\modelname{} has $18.63$M more online parameters but reduces model-only
inference time by $25.2\%$ and uses similar GPU memory. However, loading and
transferring the cached visual memories adds substantial I/O overhead, increasing
the end-to-end inference time from $16.01$ to $24.94$ ms per batch. A similar
pattern appears during training: \modelname{} reduces model computation time by
$5.7\%$, but its total step time and peak GPU memory are slightly higher than
those of GreaseLM. Overall, \modelname{} performs model computation faster, while
its current cache loading makes end-to-end inference and training
slower. Appendix~\ref{supp:parameter-composition} and~\ref{supp:computational-cost} provide detailed breakdowns of the online parameter count and comprehensive measurements of FLOPs, I/O, memory usage, and training cost.

\begin{table}[t]
\centering
\begingroup
\small
\renewcommand{\arraystretch}{0.94}
\setlength{\tabcolsep}{3.2pt}
\begin{tabular}{@{}lrrrr@{}}
\toprule
\textbf{Method} & \textbf{Online Param.} & \textbf{Compute} &
\textbf{E2E} & \textbf{CUDA} \\
& & \multicolumn{2}{c}{\textit{ms/batch}} & \textit{GiB} \\
\midrule
GreaseLM & 119.64M & 15.48 & 16.01 & 0.78 \\
\modelname{} & 138.27M & 11.57 & 24.94 & 0.76 \\
\bottomrule
\end{tabular}
\endgroup
\caption{\textbf{Online inference cost on MedQA.} ``Compute'' measures model
execution only, whereas ``E2E'' additionally includes feature loading and
host-to-device transfer. Offline KG retrieval, rendering, and visual encoding
are excluded.}
\label{tab:online-cost}
\end{table}

\section{Discussion and Limitations}
\label{sec:discussion}

Offline encoding changes where the computation is paid, but it does not remove the cost. Building $28{,}975$ candidate memories for CSQA took about three hours and
required about 21~GiB of storage. This one-time cost can be shared across training
epochs, seeds, and evaluation runs when the retrieved subgraphs remain fixed.
The online model computation is faster than GreaseLM on MedQA, but the current
one-file-per-candidate cache increases end-to-end inference time from $16.01$ to $24.94$ ms per batch. More efficient storage and loading strategies, such as packed files, memory-mapped storage, and batched cache access, could alleviate this overhead.

Fixed-resolution rendering also limits the amount of graph information that can be preserved. Dense subgraphs may lead to small text, crowded layouts, or cropped content, while our rendering analysis further shows that relation labels and layout density materially affect accuracy.

\section{Conclusion}
\label{sec:conclusion}

We presented \modelname{}, a KG-enhanced MCQA model that compiles each retrieved candidate-specific subgraph into reusable visual memory and accesses it only at the final LM layer through gated cross-attention and dual-view summarization. Across CSQA, OBQA, and MedQA, \modelname{} consistently outperforms strong LM--GNN baselines and exceeds a matched text-only control by $4.2$, $6.5$, and $5.1$ percentage points, respectively, while using approximately one eighteenth of the online parameters of the 7B GraphVis model. These results demonstrate that offline visual memory provides an effective and parameter-efficient alternative to repeatedly encoding retrieved graph evidence during online inference. Future work will focus on reducing cache I/O overhead and extending the framework to denser graphs, dynamic graph, and broader visual question-answering settings.

\bibliography{aaai2027}

\clearpage

\appendix
\setcounter{secnumdepth}{1}

\section{Additional Implementation Details}
\label{supp:implementation}

The cache stores the frozen document-encoder features $\mathcal{V}$. Layer
normalization and a two-layer projector with hidden size $4096$ map these
features to the language-model space. Both components are trained online, but
gradients stop at $\mathcal{V}$ and do not pass through the renderer or
document encoder.

We use an effective batch size of $256$ and a micro-batch size of $16$. The
projector/cross-attention dropout rates are $0.2/0.2$ for CSQA, $0.1/0.3$ for
OBQA, and $0.1/0.1$ for MedQA. All other dropout rates are $0.2$.

\section{Fusion-Depth Results}
\label{supp:fusion-depth}

Table~\ref{tab:fusion-depth-ablation} reports test accuracy for all fusion
depths $k\in\{1,\ldots,7\}$. We selected $k=1$ using validation accuracy. All
other components and training settings are fixed across rows.

\begin{table}[!ht]
    \centering
    \begingroup
    \small
    \def\kbase#1{#1\,{\small(+0.0)}}
    \def\kloss#1#2{#1\,{\small(#2)}}
    \setlength{\tabcolsep}{3pt}
    \begin{tabular}{@{}lccc@{}}
    \toprule
    \textbf{$k$} & \textbf{CSQA} & \textbf{OBQA} & \textbf{MedQA} \\
    \midrule
    \textbf{1} & \kbase{\textbf{75.2}} & \kbase{\textbf{85.6}} & \kbase{\textbf{42.7}} \\
    2 & \kloss{73.9}{$-1.3$} & \kloss{83.0}{$-2.6$} & \kloss{39.6}{$-3.1$} \\
    3 & \kloss{73.7}{$-1.5$} & \kloss{82.4}{$-3.2$} & \kloss{38.1}{$-4.6$} \\
    4 & \kloss{74.5}{$-0.7$} & \kloss{83.4}{$-2.2$} & \kloss{39.4}{$-3.3$} \\
    5 & \kloss{74.7}{$-0.5$} & \kloss{83.4}{$-2.2$} & \kloss{39.3}{$-3.4$} \\
    6 & \kloss{73.9}{$-1.3$} & \kloss{84.2}{$-1.4$} & \kloss{39.8}{$-2.9$} \\
    7 & \kloss{74.1}{$-1.1$} & \kloss{84.4}{$-1.2$} & \kloss{40.8}{$-1.9$} \\
    \bottomrule
    \end{tabular}
    \endgroup
    \caption{Test accuracy (\%) across fusion depths. Parentheses show the
    percentage-point change from the validation-selected depth $k=1$.}
    \label{tab:fusion-depth-ablation}
\end{table}

\section{DVMS Readout and Query-Budget Results}
\label{supp:dvms-ablations}

Table~\ref{tab:dvms-ablation} reports the readout and query-budget
results. Here, $G$ and $L$ denote the numbers of learnable global and local
queries. Panel~(a) replaces DVMS attention with mean pooling. The corresponding
row in Table~\ref{tab:component-ablations} uses the validation-selected budget
for each dataset: $(4,8)$ for CSQA and MedQA and $(1,8)$ for OBQA. Its three
values are therefore $73.9$, $83.6$, and $40.0$. Panel~(b) varies both DVMS
query budgets. We selected the bold configurations using validation accuracy;
the table reports their test accuracy and the full test grid.

\begin{table}[!ht]
    \centering
    \begingroup
    \small
    \renewcommand{\arraystretch}{0.92}
    \textbf{(a) Mean-pooling readout.}
    \vspace{0.1em}

    \setlength{\tabcolsep}{3.2pt}
    \begin{tabular}{@{}lccccc@{}}
    \toprule
    \textbf{Method} & \textbf{$G$} & \textbf{$L$} & \textbf{CSQA} & \textbf{OBQA} & \textbf{MedQA} \\
    \midrule
    Mean pooling & 1 & 1 & 74.0 & 83.4 & 38.5 \\
    Mean pooling & 1 & 8 & 74.1 & 83.6 & 37.5 \\
    Mean pooling & 2 & 16 & 73.7 & 83.8 & 40.1 \\
    Mean pooling & 4 & 8 & 73.9 & 82.8 & 40.0 \\
    Mean pooling & 8 & 16 & 74.3 & 83.4 & 39.2 \\
    \bottomrule
    \end{tabular}
    \vspace{0.25em}

    \textbf{(b) DVMS query budgets.}
    \vspace{0.1em}

    \setlength{\tabcolsep}{4pt}
    \begin{tabular}{@{}ccccc@{}}
    \toprule
    \textbf{$G$} & \textbf{$L$} & \textbf{CSQA} & \textbf{OBQA} & \textbf{MedQA} \\
    \midrule
    1 & 1  & 73.2 & 82.0 & 41.4 \\
    1 & 2  & 73.9 & 82.6 & 39.8 \\
    1 & 4  & 73.2 & 83.1 & 40.9 \\
    1 & 8  & 73.5 & \textbf{85.6} & 39.9 \\
    1 & 16 & 73.9 & 81.2 & 40.1 \\
    2 & 1  & 73.3 & 82.0 & 39.7 \\
    2 & 2  & 73.6 & 82.6 & 39.2 \\
    2 & 4  & 73.7 & 81.2 & 39.9 \\
    2 & 8  & 73.3 & 83.8 & 39.8 \\
    2 & 16 & 73.2 & 84.6 & 40.8 \\
    4 & 1  & 73.7 & 82.0 & 40.2 \\
    4 & 2  & 73.9 & 82.4 & 40.5 \\
    4 & 4  & 73.7 & 82.1 & 40.6 \\
    4 & 8  & \textbf{75.2} & 84.5 & \textbf{42.7} \\
    4 & 16 & 74.6 & 83.1 & 39.2 \\
    8 & 1  & 73.0 & 83.6 & 39.9 \\
    8 & 2  & 75.2 & 81.6 & 40.8 \\
    8 & 4  & 74.3 & 82.1 & 40.6 \\
    8 & 8  & 73.6 & 83.4 & 39.4 \\
    8 & 16 & 74.9 & 83.3 & 41.0 \\
    \bottomrule
    \end{tabular}
    \endgroup
    \caption{Test accuracy (\%) for the readout ablations. Panel~(a) uses mean
    pooling, and panel~(b) uses DVMS. $G$ and $L$ are the global and local query
    budgets. Bold values indicate the best result for each dataset.}
    \label{tab:dvms-ablation}
\end{table}

\section{Graph-Corruption Results}
\label{supp:graph-corruption}

\begin{figure*}[t]
    \centering
    \includegraphics[width=0.75\textwidth]{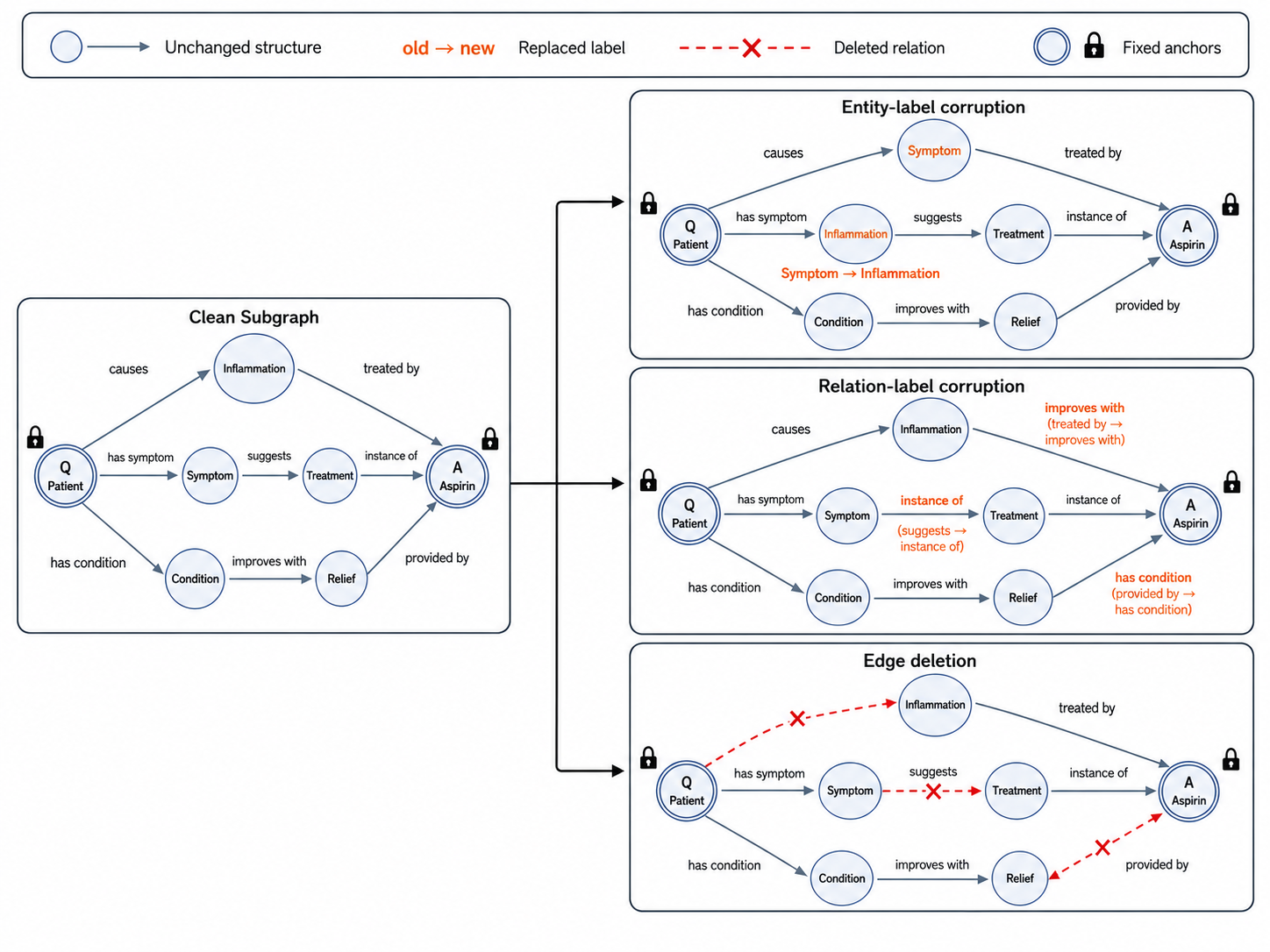}
    \caption{Graph-corruption examples: replacing relation labels, deleting
    edges, and replacing the names of non-seed entities.}
    \label{fig:graph-noise}
\end{figure*}

Table~\ref{tab:robustness} adds the $25\%$, $50\%$, and $75\%$ corruption levels
to the endpoints in Table~\ref{tab:corruption-endpoints}. For each entry, we
train and test the model with the same corruption operator and strength. These
results therefore measure matched train--test corruption, not the test-time
robustness of a model trained on clean graphs. Figure~\ref{fig:graph-noise}
illustrates the three corruption operators.

\begin{table}[!ht]
    \centering
    \begingroup
    \small
    \renewcommand{\arraystretch}{0.94}
    \setlength{\tabcolsep}{3.4pt}
    \textbf{(a) Relation-label replacement.}
    \vspace{0.1em}

    \begin{tabular}{@{}lccccc@{}}
    \toprule
    \textbf{Model} & \textbf{0\%} & \textbf{25\%} & \textbf{50\%} & \textbf{75\%} & \textbf{100\%} \\
    \midrule
    GreaseLM & 38.41 & 38.33 & 38.33 & 38.33 & 38.49 \\
    \textbf{\modelname{}} & \textbf{42.66} & \textbf{42.34} & \textbf{42.58} & \textbf{42.73} & \textbf{42.34} \\
    \bottomrule
    \end{tabular}
    \vspace{0.25em}

    \textbf{(b) Edge deletion.}
    \vspace{0.1em}

    \begin{tabular}{@{}lccccc@{}}
    \toprule
    \textbf{Model} & \textbf{0\%} & \textbf{25\%} & \textbf{50\%} & \textbf{75\%} & \textbf{100\%} \\
    \midrule
    GreaseLM & 38.41 & 38.33 & 38.10 & 38.10 & 38.18 \\
    \textbf{\modelname{}} & \textbf{42.66} & \textbf{42.21} & \textbf{42.97} & \textbf{42.89} & \textbf{42.58} \\
    \bottomrule
    \end{tabular}
    \vspace{0.25em}

    \textbf{(c) Entity-name replacement.}
    \vspace{0.1em}

    \begin{tabular}{@{}lccccc@{}}
    \toprule
    \textbf{Model} & \textbf{0\%} & \textbf{25\%} & \textbf{50\%} & \textbf{75\%} & \textbf{100\%} \\
    \midrule
    GreaseLM & 38.41 & 38.41 & 38.41 & 38.41 & 38.41 \\
    \textbf{\modelname{}} & \textbf{42.66} & \textbf{42.58} & \textbf{42.34} & \textbf{42.63} & \textbf{42.41} \\
    \bottomrule
    \end{tabular}
    \endgroup
    \caption{MedQA-USMLE test accuracy (\%) under matched train--test graph
    corruption. Each model is retrained for every operator and strength.}
    \label{tab:robustness}
\end{table}

\section{Dataset and Cache Statistics}
\label{supp:dataset-statistics}

Table~\ref{tab:supp-dataset-statistics}(a) compares the official CSQA splits
with the in-house (IH) evaluation splits used by KagNet and GreaseLM
\cite{lin2019kagnet,zhang2022greaselm}. Because the official test labels are not
public, all reported CSQA results use the IH protocol. It divides the $9{,}741$
labeled training questions into $8{,}500$ IHtrain and $1{,}241$ IHtest examples
and uses the official $1{,}221$-example development set as IHdev.

Panel~(b) reports the preprocessing volume. \modelname{} creates one image and
one cached memory per answer choice: five per CSQA question and four per OBQA
or MedQA question. We also preprocessed the official CSQA test set, although it
was not used for evaluation.

\begin{table}[!ht]
    \centering
    \begingroup
    \small
    \renewcommand{\arraystretch}{0.95}
    \textbf{(a) Dataset split sizes.}
    \vspace{0.1em}

    \setlength{\tabcolsep}{2.7pt}
    \begin{tabular}{@{}llrrr@{}}
        \toprule
        \textbf{Dataset} & \textbf{Protocol} & \textbf{Train} & \textbf{Dev} & \textbf{Test} \\
        \midrule
        CSQA & Official & 9,741 & 1,221 & 1,140 \\
        CSQA & IH evaluation & 8,500 & 1,221 & 1,241 \\
        OBQA & Standard & 4,957 & 500 & 500 \\
        MedQA & Standard & 10,178 & 1,272 & 1,273 \\
        \bottomrule
    \end{tabular}
    \vspace{0.25em}

    \textbf{(b) Candidate-level image counts.}
    \vspace{0.1em}

    \setlength{\tabcolsep}{3.5pt}
    \begin{tabular}{@{}llrrr@{}}
        \toprule
        \textbf{Dataset} & \textbf{Protocol} & \textbf{Choices} & \textbf{Questions} & \textbf{Pairs} \\
        \midrule
        CSQA & Official & 5 & 12,102 & 60,510 \\
        CSQA & IH evaluation & 5 & 10,962 & 54,810 \\
        OBQA & Standard & 4 & 5,957 & 23,828 \\
        MedQA & Standard & 4 & 12,723 & 50,892 \\
        \bottomrule
    \end{tabular}
    \endgroup
    \caption{Dataset split sizes and candidate-level image counts. Reported
    CSQA results use the IH evaluation protocol.}
    \label{tab:supp-dataset-statistics}
\end{table}

\section{Online Parameter Counts}
\label{supp:parameter-composition}

Table~\ref{tab:model-parameter-composition} divides the online parameters into
the text backbone and evidence interface. We count deduplicated named
parameters in the loaded checkpoints. For \modelname{}, the evidence interface
contains layer normalization, the visual-memory projector, gated
cross-attention, DVMS, and the prediction head. The count excludes the frozen
$\sim$580M-parameter \deepenc{} V2 encoder, which is used only to build the
cache. The GreaseLM count also excludes its frozen external ConceptNet entity
table.

\begin{table}[!ht]
    \centering
    \begingroup
    \small
    \setlength{\tabcolsep}{2.5pt}
    \begin{tabular}{lrr}
        \toprule
        \textbf{Component} & \textbf{GreaseLM} & \textbf{\modelname{}} \\
        \midrule
        \multicolumn{3}{l}{\textbf{(a) SapBERT-Base (MedQA)}} \\
        \addlinespace
        Text backbone & 109.48 & 109.48 \\
        Evidence interface + head & 10.16 & 28.79 \\
        \midrule
        \textbf{Total} & \textbf{119.64} & \textbf{138.27} \\
        \midrule
        \multicolumn{3}{l}{\textbf{(b) RoBERTa-large (CSQA/OBQA)}} \\
        \addlinespace
        Text backbone & 355.36 & 355.36 \\
        Evidence interface + head & 3.62 & 46.74 \\
        \midrule
        \textbf{Total} & \textbf{358.98} & \textbf{402.10} \\
        \bottomrule
    \end{tabular}
    \caption{Online parameter counts (millions) with matched text backbones.
    Counts exclude the offline visual encoder and GreaseLM's frozen external
    ConceptNet entity table.}
    \label{tab:model-parameter-composition}
    \endgroup
\end{table}

For the first four MedQA epochs, SapBERT is frozen and \modelname{} has
$28{,}790{,}275$ trainable parameters. After unfreezing SapBERT, all
$138{,}272{,}515$ online parameters are trainable. These are two training
phases of the same model.

\section{Computational Cost}
\label{supp:computational-cost}

Table~\ref{tab:computational-cost} reports online cost on MedQA. Both models
use SapBERT-Base, a maximum text
length of $512$, and batches containing one four-choice question. Measurements
use one NVIDIA H100 PCIe GPU, PyTorch $2.4.1$, CUDA $12.1$, bfloat16 mixed
precision, and FP32 master parameters. The data loader uses one process, and
the in-memory feature cache is disabled to include storage access. Retrieval,
rendering, and document encoding are offline and are therefore excluded.

\begin{table}[!ht]
    \centering
    \begingroup
    \small
    \renewcommand{\arraystretch}{0.94}
    \textbf{(a) Inference (batch size 1).}
    \vspace{0.1em}

    \setlength{\tabcolsep}{2.6pt}
    \begin{tabular}{@{}lrrrrr@{}}
        \toprule
        \textbf{Method} & \textbf{Params.} & \textbf{GFLOPs/q} &
        \textbf{Forward} & \textbf{Input} & \textbf{E2E} \\
        & & & \multicolumn{3}{c}{\textit{ms/batch}} \\
        \midrule
        GreaseLM & 119.64M & 352.97 & 15.48 & 0.48 & 16.01 \\
        \modelname{} & 138.27M & 456.88 & 11.57 & 13.31 & 24.94 \\
        \bottomrule
    \end{tabular}
    \vspace{0.3em}

    \textbf{(b) Unfrozen training (effective/micro-batch size $1/1$).}
    \vspace{0.1em}

    \setlength{\tabcolsep}{2.7pt}
    \begin{tabular}{@{}lrrrrr@{}}
        \toprule
        \textbf{Method} & \textbf{Trainable} & \textbf{Compute} &
        \textbf{I/O+H2D} & \textbf{Total} & \textbf{q/s} \\
        & \textit{params.} & \multicolumn{3}{c}{\textit{ms/step}} & \\
        \midrule
        GreaseLM & 119.64M & 141.41 & 1.07 & 142.48 & 7.07 \\
        \modelname{} & 138.27M & 133.32 & 18.59 & 151.90 & 7.50 \\
        \bottomrule
    \end{tabular}
    \vspace{0.3em}

    \textbf{(c) Peak memory (GiB).}
    \vspace{0.1em}

    \setlength{\tabcolsep}{4pt}
    \begin{tabular}{@{}lrrrr@{}}
        \toprule
        \textbf{Method} & \textbf{Inf. CUDA} & \textbf{Inf. RSS} &
        \textbf{Train CUDA} & \textbf{Train RSS} \\
        \midrule
        GreaseLM & 0.78 & 3.48 & 2.44 & 3.62 \\
        \modelname{} & 0.76 & 4.00 & 3.01 & 4.09 \\
        \bottomrule
    \end{tabular}
    \endgroup
    \caption{MedQA-USMLE online cost. Input time includes storage access and
    host-to-device transfer. In panel~(b), q/s is compute-only throughput.}
    \label{tab:computational-cost}
\end{table}

The FLOP profiler reports $456.88$ GFLOPs per question for \modelname{} and
$352.97$ for GreaseLM. It may omit sparse or custom graph operations, so these
values are not a complete operation-level comparison. Panel~(c) reports both
CUDA allocation and resident set size (RSS).

Using the measured end-to-end rates, we estimate the runtime of each full
training recipe. GreaseLM uses effective/micro-batch sizes of $128/2$ with
RAdam; \modelname{} uses $256/16$ with AdamW and freezes SapBERT for four
epochs. On the $10{,}178$-question MedQA training split, one GreaseLM epoch
takes about $6.66$ minutes, while one frozen or unfrozen \modelname{} epoch
takes about $4.37$ or $5.06$ minutes. The full schedules---$17$ GreaseLM epochs
and $4+11$ \modelname{} epochs---take an estimated $1.89$ and $1.22$ hours,
respectively. Because the two models use different optimization settings,
these estimates compare complete training recipes rather than architectures
alone.

\end{document}